\documentclass{article}

\PassOptionsToPackage{numbers,compress}{natbib}

\usepackage[final]{tackling_climate_workshop_style}

\usepackage[utf8]{inputenc}   %
\usepackage[T1]{fontenc}      %
\usepackage{hyperref}         %
\usepackage{url}              %
\usepackage{booktabs}         %
\usepackage{amsfonts}         %
\usepackage{amsmath}
\usepackage{amssymb}
\usepackage{nicefrac}         %
\usepackage{microtype}        %
\usepackage{graphicx}
\usepackage{float}            %
\usepackage{xcolor}
\usepackage{subcaption}       %
\usepackage{siunitx}
\usepackage{verbatim}

\hypersetup{colorlinks=true, linkcolor=black, citecolor=black, urlcolor=blue}

\title{Global Solar Siting from Geospatial Embeddings}

\author{%
  Abdullah Rehman$^{1,2,*}$ \qquad Quentin Paletta$^{3,4}$ \qquad Rui Song$^{2}$ \\[4pt]
  \\
  $^{1}$Department of Engineering, King's College London, London, UK \\
  $^{2}$Department of Physics, University of Oxford, Oxford, UK \\
  $^{3}$ESA $\Phi$-Lab, ESRIN, European Space Agency, Frascati, Italy \\[2pt]
  $^{4}$ESA Climate Office, ECSAT, European Space Agency, Harwell, UK \\[2pt]
  \\
  $^{*}$\texttt{abdullah.rehman@kcl.ac.uk}
}

\begin{document}
\maketitle

\begin{abstract}
Global solar photovoltaic (PV) capacity is rapidly growing, helping decrease our emissions and dependence on fossil fuels. However, optimally siting a solar plant remains challenging, requiring anticipation of geographical, urban, and infrastructure needs. Current methods rely on multi-criteria decision analysis (MCDA) using hand-picked geophysical layers and solar-resource products, however these strategies are not geographically transferable and do not capture the practical requirements of solar sites. We use geospatial embeddings, dense vectors representing geophysical properties of a location, to analyse \num{38591} existing solar sites and the land they were built on to learn the latent solar suitability signal. We assess a set of linear and non-linear models and select a soft-voting ensemble of a random forest and a multilayer perceptron (RF+MLP ensemble), chosen for accuracy and large scale deployability. Validated against future installations in thirteen countries, the ensemble separates existing solar sites from random land with a median ROC-AUC of \num{0.90}. Compared with published MCDA strategies, the ensemble increases ROC-AUC by \num{0.06}--\num{0.30} in every country. The ranking also concentrates real deployment: the most-suitable \num{20}\% of land holds \num{84}\% of estimated generation (annual energy yield) and \num{67}\% of installed capacity (potential power output). After applying standard siting constraints, \num{5.6}\% of global land (\num{9.8}\% if cropland is permitted) remains highly suited to solar. The global \SI{10}{\meter} resolution solar suitability dataset and code can be found here: \texttt{https://github.com/Abdullahr29/Global-Solar-Siting}.

\end{abstract}

\section{Introduction}
\label{sec:intro}

Energy production and consumption accounts for approximately \num{76}\% of global greenhouse gas emissions \cite{wri}, primarily due to the burning and consumption of fossil fuels \cite{noauthor_greenhouse_nodate}. Meeting net-zero targets requires renewable energy sources to displace fossil fuel driven generation at an unprecedented pace \cite{kannan_solar_2016}. Solar photovoltaic (PV) is a key technology in this energy transition, accounting for \num{81}\% of all new renewable capacity added in 2024, and reaching \num{3} terawatt (TW) in installed capacity globally in 2026 \cite{global_solar_2025}. By 2030, the world is projected to be installing over \num{1} TW of solar annually, and as deployment accelerates and highly accessible locations are developed, the binding constraint is increasingly in where to build and identifying land that will make a successful site \cite{global_solar_2025, gorman_grid_2025}. 

Solar siting currently requires the analysis of several geospatial features pertaining to topography, land cover, solar resources, and urban infrastructure \cite{alhammad_optimal_2022}. These layers are used in Multi-Criteria Decision Analysis (MCDA) frameworks to optimise solar placement, but these approaches are highly variable, with decisions driven by feature weighting and often limited to one country or region \cite{kozlov_challenges_2021, matulaitis_multi-criteria_2016, mathebula_application_2025, richards_sustainable_2025, chen_jong_multi-criteria_2024}. Machine learning (ML) approaches in solar research have primarily focused on forecasting solar energy generation \cite{sebastianelli_machine_2024, reddy_solar_2025}. Studies which apply ML to solar siting follow a similar pattern to MCDA, engineering geospatial features for classical models, with decision tree ensembles often performing well but still being limited to one region \cite{shorabeh_decision_2022, sun_spatial_2023, chen_solar_2026, tao_spatial_2025}.

Where global products do exist they model PV power potential \cite{noauthor_global_nodate, saxena_modelling_2023} or map existing installations, such as the Global Solar Power Tracker \cite{noauthor_global_nodate-1}. Solar siting reflects not only solar-resource availability but also policy, infrastructure and land-use constraints \cite{song_coal_2026}. A forward-looking global solar suitability layer would supplement these products and support faster, better-justified siting.

Learning the inherent property of land which is suitable for solar through a set of predefined features is challenging. Recent advancements in geospatial representation learning offer a new pathway to tackling this problem \cite{klemmer_earth_2026}. Global embeddings such as AlphaEarth Foundations (AEF) encode a diverse mix of geophysical and meteorological properties alongside remote sensing information into a high dimensional vector \cite{brown_alphaearth_2025}, and even simpler models trained on embeddings can transfer land-cover labels across continents \cite{houriez_scalable_nodate}. While geospatial embeddings are not as interpretable as data from optical sensors such as Sentinel-2, they circumvent many of the challenges that come with using raw imagery, including cloud masking, seasonal adjustments, and processing differences, challenges faced in solar infrastructure detection \cite{zeng_global_2025, song_coal_2026}. Embeddings instead provide an analysis-ready, gap-filled annual summary of every \SI{10}{\meter} pixel, learnt from many sensors at once \cite{brown_alphaearth_2025}.

\textbf{Aims and Objectives.} In this study, we explore whether geospatial embeddings can encode a non-obvious, future-looking property of land: the probability and suitability of future solar development. We evaluate a range of models trained on geospatial embeddings of existing solar sites and non-solar land, validating their performance on future solar installations across countries, and comparing them against published MCDA approaches and existing solar-resource products. Finally, we release a global \SI{10}{\meter} solar suitability layer, empowering future solar siting and the renewable energy transition.

\section{Data}
\label{sec:data}

\begin{figure}[tbp]
  \centering
  \includegraphics[width=\textwidth]{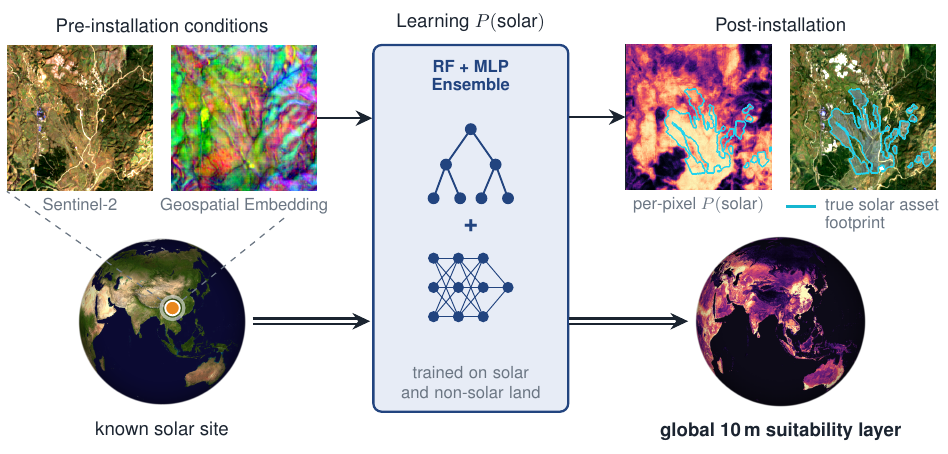}
  \caption{The pipeline for learning solar suitability. For a known solar site, the per-pixel geospatial embedding of the land (here the 64-dimensional AEF embedding) is sampled two years before installation, capturing the pre-installation conditions. A random forest and multilayer perceptron ensemble, trained on solar and globally-random non-solar land, maps each embedding pixel to $P(\text{solar})$. Post-installation, the true solar asset footprint (outlined) shows the developed site. \mbox{Sentinel-2} imagery is shown for visual reference only and is not a model input. Applied globally, the model yields an openly-released global 10\,m suitability layer.}
  \label{fig:pipeline}
\end{figure}

\textbf{Positive sites.} From the Global PV Facility (GPVF) inventory of \num{140945} facilities (with delineated footprints and installation dates) we draw a country-balanced sample of \num{38591} sites \cite{song_global_2026}. 64-dimensional AEF embeddings of the sites 2 years before solar installation are then used for training; the AEF embedding record begins in 2017, therefore we can only study sites from 2019 to 2024. The sample spans \num{154} countries. Each site's delineated footprint is rasterised onto the \SI{10}{\meter} embedding grid and the pixels inside it form the positive labels. The task is to anticipate development rather than to recognise existing panels, as shown in Fig. \ref{fig:pipeline}. 

\textbf{Negative sites.} Non-solar land is sampled as globally-random parcels at least \SI{5}{km} from any solar site of any year, present in either the GPVF or the TZ-SAM, a secondary global inventory of solar energy facilities \cite{phillpott_solar_2024}, datasets. AEF embeddings for negative sites are sampled randomly from any year between 2017-2022. The final solar-to-non-solar ratio is ${\sim}1{:}5$.

\textbf{Testing \& validation.} Scenes are split \num{80}/\num{20} into train and test as whole chips, to minimise pixel leakage (see Appendix \ref{app:train} for exact values). We additionally evaluate the models against the TZ-SAM inventory, testing specifically against sites missing from the GPVF dataset, our training database. 

\section{Methodology}
\label{sec:method}

\textbf{Model benchmarking.} To assess the optimal mechanism of extracting solar suitability from the AEF embedding, we compare nine models of increasing complexity: a naive mean-embedding prototype scored by cosine similarity \cite{yuan_review_2026}; a quadratic discriminant (QDA) \cite{hastie_elements_2009}; a logistic probe \cite{corley_pixels_2026}; MaxEnt, the standard species-distribution and GIS-suitability baseline \citep{phillips2006maxent}; random Fourier features with a logistic head \cite{rahimi_random_2007}; a random forest (RF) \citep{breiman2001random}; histogram gradient boosting \cite{ke_lightgbm_2017}; a multilayer perceptron (MLP) \cite{rumelhart_learning_1986}; and a soft-voting ensemble of the RF and MLP (RF+MLP ensemble) \cite{kuncheva_combining_2004}. All nine are trained on a common balanced sample of training pixels; full formulations and hyperparameters are given in Appendix~\ref{app:methods}.

\textbf{Validation.} On held-out pixels we report the area under the Receiver Operating Characteristic curve (ROC-AUC), the probability that a randomly chosen solar pixel is ranked above a randomly chosen non-solar pixel, and the area under the precision--recall curve (PR-AUC, equivalently average precision), which is more informative under the ${\sim}1{:}5$ class imbalance. We further validate the RF+MLP ensemble forward in time, across 13 held-out countries and under a leak-proof temporal split, cross-checking every country against the independent TZ-SAM inventory; the full protocol, chip-level and per-country results, and leakage ablations are given in Appendix \ref{app:examples} and \ref{app:validation}. 

\textbf{Global-scale analysis.} We investigate what the suitability ranking implies for real deployment. Joining each validation site to its TZ-SAM capacity estimate, we measure how the world's installed capacity distributes across the ranked land. Joining instead to the GPVF modelled annual generation, we ask whether the ranking concentrates estimated generation, testing the correlation between suitability and energy yield. Finally, we threshold the suitability map at the tenth percentile of real-site scores, and impose literature-derived siting cut-offs (slope $\leq5^\circ$, grid distance $\leq\SI{10}{km}$, protected-area exclusion) to report a grid-constrained realistic fraction of global land highly suited for solar installation.  

\textbf{Comparison to MCDA.} We benchmark the ensemble model against the MCDA that underpins conventional siting. Using two sets of published expert weights (\citet{richards_sustainable_2025} and \citet{chen_jong_multi-criteria_2024}) and two standard aggregators (weighted linear combination and TOPSIS), we construct suitability maps from solar-relevant features and score each MCDA approach on the identical site-versus-random task against the ensemble. MCDA model explanations, weights, and parameters are given in Appendix \ref{app:mcda}. Finally, we conduct a covariate analysis to quantify the extent to which the ensemble's suitability signal is explained by standard GIS layers, including land cover, slope, irradiance, and solar resource products such as PVOUT (see Appendix \ref{app:covariate}).

\section{Results}
\label{sec:results}

\textbf{The siting signal is real but strongly non-linear.} Table~\ref{tab:benchmark} ranks the extractors on held-out pixels and on held-out sites per country. The naive mean-embedding prototype---the classic embedding similarity search mechanism---separates solar from non-solar land only weakly (ROC-AUC \num{0.70}), so suitability is not a simple direction in embedding space. Linear and kernel models recover most of the signal (logistic and MaxEnt \num{0.87}, random Fourier features \num{0.89}) and the non-linear models reach \num{0.92}. Averaging a random forest and an MLP trained on the same sample improves on every single model across every validation test, and this ensemble is therefore chosen for deployment. The released global layer is the same RF+MLP ensemble refit on all data.

\begin{table}[t]
  \centering
  \caption{Extraction-method benchmark. Pixel: held-out test pixels (PR-AUC chance rate \num{0.17}). Country: ROC-AUC of future sites against random national land, averaged over the thirteen validation countries weighted by site count. TZ-SAM: independent sites absent from our training inventory. All methods are trained on one common balanced sample; combining a random forest and an MLP into an ensemble improves on every single model in the benchmark.}
  \label{tab:benchmark}
  \small
  \setlength{\tabcolsep}{4pt}
  \begin{tabular}{@{}l l r r r r@{}}
    \toprule
    & & \multicolumn{2}{c}{Pixel} & Country & TZ-SAM \\
    \cmidrule(lr){3-4}
    Method & Family & ROC-AUC & PR-AUC & ROC-AUC & ROC-AUC \\
    \midrule
    Mean-embedding cosine & prototype & 0.702 & 0.318 & 0.751 & 0.739 \\
    QDA & discriminant & 0.854 & 0.518 & 0.830 & 0.834 \\
    Logistic probe & linear & 0.866 & 0.551 & 0.828 & 0.823 \\
    MaxEnt & log-linear (SDM) & 0.874 & 0.577 & 0.833 & 0.827 \\
    Random Fourier features & kernel & 0.891 & 0.628 & 0.856 & 0.841 \\
    Random forest & tree ensemble & 0.914 & 0.696 & 0.893 & 0.874 \\
    Hist.\ gradient boosting & boosted trees & 0.919 & 0.713 & 0.896 & 0.880 \\
    MLP & neural & 0.924 & 0.728 & 0.893 & 0.872 \\
    \textbf{RF+MLP ensemble} & ensemble & \textbf{0.932} & \textbf{0.752} & \textbf{0.904} & \textbf{0.882} \\
    \bottomrule
  \end{tabular}
\end{table}

\textbf{The model anticipates future solar.} Trained only on land as it appeared at least two years before installation, the model separates real future sites from random national land in all thirteen countries, with forward ROC-AUC spanning \num{0.83}--\num{0.95} (see Appendix \ref{app:validation}). As Fig.~\ref{fig:rf} demonstrates: at site scale the model scores the pre-installation embedding of land that later became solar as highly suitable (Fig.~\ref{fig:rf}a), and at national scale (Fig.~\ref{fig:rf}b, Greece) the \num{876} validation sites concentrate on the suitable minority, their scores cleanly separated from random land (inset histogram, ROC-AUC \num{0.894}).

\begin{figure}[b]
  \centering
  \includegraphics[width=0.82\textwidth]{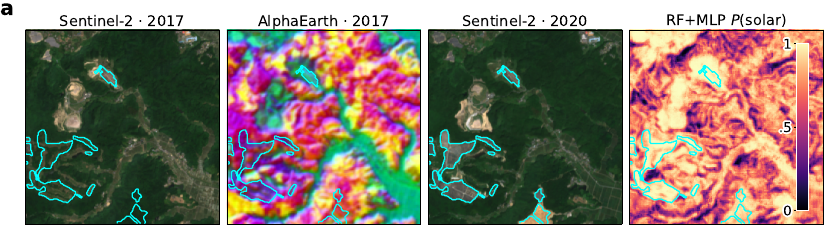}\\[3pt]
  \includegraphics[width=0.82\textwidth]{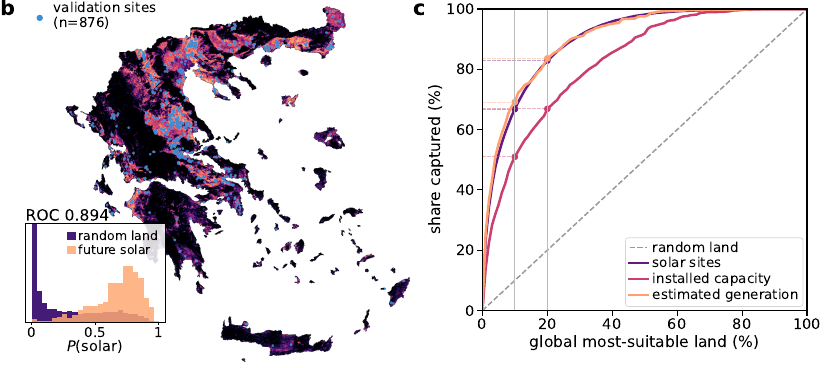}
  \caption{\textbf{(a)}~A solar site (Japan) with footprints outlined: Sentinel-2 before construction (2017), the AEF embedding two years before installation (2017), Sentinel-2 after construction (2020), and the per-pixel RF+MLP suitability map; Sentinel-2 is shown for visual reference only. The land is forested hills and valleys in 2017, yet the pre-installation embedding scores them accurately. \textbf{(b)}~National suitability for Greece (AEF-2019) with \num{876} validation sites concentrated on the most-suitable land: the score histogram shows them cleanly separated from random national land (ROC-AUC \num{0.894}). \textbf{(c)}~Cumulative share of solar sites, installed capacity, and estimated generation captured versus the most-suitable land ranked by the ensemble; capacity concentrates less than sites and generation, as the largest farms are atypically sited (see Appendix \ref{app:size}).}
  \label{fig:rf}
\end{figure}

\textbf{The map concentrates capacity and generation.} The most-suitable \num{20}\% of land carries \num{67}\% of the world's installed capacity and \num{84}\% of its estimated generation (Fig.~\ref{fig:rf}c), and weighting validation sites by their generation lowers discrimination only slightly (ROC-AUC \num{0.93} to \num{0.90}), as the largest plants are less typically sited. Globally, \num{22.1}\% of land clears the suitability threshold and \num{5.6}\% (\num{9.8}\% if cropland is allowed) survives the grid-and-terrain constraints, so suitable land is not what limits solar, economics and policy are; the map is best used to prioritise within that abundant land.

\begin{figure}[H]
  \centering\includegraphics[width=\textwidth]{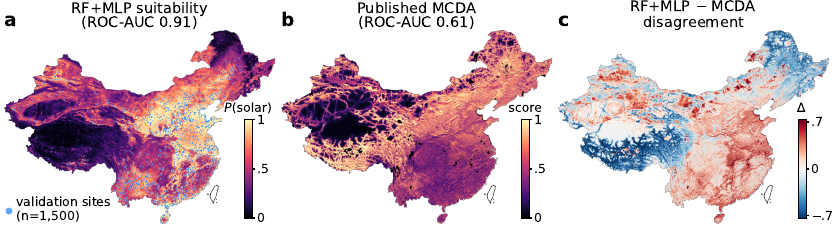}
  \caption{Both maps are shown as within-country percentiles (the share of China's land each cell outscores). \textbf{(a)}~RF+MLP suitability (AEF-2019) with future solar sites (blue; a subsample of \num{1500} of the \num{27158} scored sites): solar is built in the east, which the ensemble ranks high (ROC-AUC \num{0.91}). \textbf{(b)}~Published MCDA instead follows infrastructure and solar-resource heuristics, ranking western and northern land suitable (ROC-AUC \num{0.61}, the best of the \num{16} tested MCDA configurations). \textbf{(c)}~The difference between the RF+MLP ensemble and MCDA. Red indicates a high ensemble and low MCDA score (blue vice-versa), outlining significant disagreement in high solar regions.}
  \label{fig:china}
\end{figure}

\textbf{The model surpasses existing siting methods.} The ensemble outperforms published MCDA in every one of the thirteen countries, by \num{0.06}--\num{0.30} ROC-AUC, every gap statistically significant (Appendix~\ref{app:mcda}, Table~\ref{tab:mcda}). Fig.~\ref{fig:china} shows China, where MCDA ranks the empty western interior suitable, yet real solar is built in the east, which the ensemble correctly identifies. Regressing the ensemble's score on standard GIS layers (Appendix~\ref{app:covariate}), land cover and slope are the most important predictors in most countries, while irradiance and the long-term PV output potential (PVOUT) rarely are. PVOUT alone cannot separate sites from random land (ROC-AUC \num{0.49}). Between \num{20}\% and \num{53}\% of the ensemble score's variance is not explained by these geospatial layers, information the embedding carries beyond hand-picked features.

\section{Discussion}
\label{sec:discussion}

Our work shows that geospatial embeddings can be probed for non-obvious, forward-looking properties of land, and that solar suitability is a strongly non-linear but latent property, best extracted by combining complementary models. Future extensions of this work include the investigation of other geospatial embedding products \cite{feng_tessera_2026} and spatial models which look beyond individual pixels and consider site suitability across scales. 

Our current approach is built on the assumption that existing solar facilities are well sited, however this may limit the model's ability to identify land that is suitable in underdeveloped regions. Future approaches applying positive-unlabelled learning may mitigate this limitation, by treating unlabelled land as a mixture of suitable and unsuitable rather than as negatives. Conversely, land which has been identified as highly suitable may not remain so in the future due to climate change. Risks to solar power are emerging, to solar irradiance (wildfire smoke) \cite{corwin_solar_2025} or to solar assets themselves (flooding) \cite{IBRAHIM2024113901}, which the current model cannot foresee. This necessitates the development of future solar suitability models which incorporate climate change projections and risk assessment.

Access to solar power presents an opportunity for clean and affordable energy in developed and developing nations alike. An open, global and validated suitability layer can help governments, grid and land-use planners, and development organisations prioritise where to investigate, including in data-poor countries. Used in conjunction with environmental assessment and community consultation, the suitability layer can serve as an additional resource for data-informed siting of future solar plants with minimal environmental and social impact. 

\clearpage
\begin{ack}
This work was funded by the European Space Agency (ESA) Dragon Cooperation Programme through the project FUCEO: Exploring synergies between Chinese and European EO mission using data fusion (funding ID 95452). This work used JASMIN, the UK's collaborative data analysis environment (\url{https://www.jasmin.ac.uk}), and Google Earth Engine for data access and processing. The authors thank the ESA $\Phi$--lab and ESA Climate Office for
their support during this research.

\end{ack}

\bibliographystyle{unsrtnat}
\bibliography{references}

\clearpage
\appendix

\section{Training Data Size}
\label{app:train}
Table~\ref{tab:data} summarises the training data. Positive pixels are the rasterised footprints of the country-balanced solar-site sample; negative pixels are drawn from the globally-random non-solar parcels (Section~\ref{sec:data}). The number of training pixels and positive-to-negative ratio were tested against model performance, with the current $\sim$1:5 solar-to-non-solar pixel ratio performing optimally. The \num{80}/\num{20} train/test split is made over whole scenes, so no two chips from the same site fall on opposite sides of the split.

\begin{table}[H]
  \centering
  \caption{ Distribution of training data by chips and pixels. Scenes are split \num{80}/\num{20} into train and test as whole chips. }
  \label{tab:data}
  \small
  \setlength{\tabcolsep}{5pt}
  \begin{tabular}{@{}l p{4.2cm} r r@{}}
    \toprule
    Class & Source & Chips (train\,:\,test) & Pixels (train\,:\,test) \\
    \midrule
    Positive (solar) & PV-facility inventory; footprint masks
     & \num{65811}\,:\,\num{16337} & \num{6193548}\,:\,\num{1550685} \\
    Negative (non-solar) & Random parcels $\geq$\SI{5}{km} from any solar site; TZ-SAM-screened
     & \num{10775}\,:\,\num{2665} & \num{30170000}\,:\,\num{7462000} \\
    \midrule
    Total & & \num{76586}\,:\,\num{19002} & \num{36363548}\,:\,\num{9012685} \\
    \bottomrule
  \end{tabular}
\end{table}

\section{Extraction methods: formulations and parameters}
\label{app:methods}
All methods take the 64-dimensional AEF pixel embedding
$\mathbf{x}\in\mathbb{R}^{64}$ as input. The benchmark models (Table~\ref{tab:benchmark}) are trained on a common balanced subsample of \num{1400000} training pixels. Models that benefit from standardisation are fitted on features centred and scaled by a \texttt{StandardScaler} learnt on the subsample. We use the scikit-learn implementations \cite{pedregosa2011scikit} (unlisted arguments take library defaults), and PyTorch for the tuned MLP.

\textbf{Mean-embedding prototype (cosine).} A prototype $\boldsymbol{\mu}$ is the
mean of up to \num{2000000} (no change in vector with more training pixels) solar-pixel embeddings, unit-normalised to
$\hat{\boldsymbol{\mu}}$; a pixel is scored by
$s(\mathbf{x})=\mathbf{x}^{\top}\hat{\boldsymbol{\mu}}$. This is the classic
embedding-similarity approach and involves no discriminative training. Because AEF
embeddings are themselves near unit-norm, normalising the prototype (not the
inputs) recovers cosine similarity.

\textbf{Logistic probe.} $\ell_2$-regularised logistic regression ($C=1$,
\texttt{max\_iter}$=300$) on standardised embeddings with balanced class
weights, the simplest discriminative linear read-out.

\textbf{Quadratic discriminant analysis (QDA).} Class-conditional Gaussians with
separate covariance matrices, yielding a quadratic decision boundary; fitted on the
raw embeddings.

\textbf{MaxEnt.} The maximum-entropy species-distribution model
\cite{phillips2006maxent}, the standard ecological/GIS suitability baseline. Where
the \texttt{elapid} implementation is available we use it directly; otherwise we use
its infinitely-weighted-logistic-regression equivalent: an $\ell_1$-penalised
logistic regression ($C=0.5$, \texttt{saga} solver, balanced weights) on linear and
quadratic embedding features $[\mathbf{x},\mathbf{x}^{2}]$.

\textbf{Random Fourier features (RFF).} A \num{512}-dimensional random Fourier
feature map \cite{rahimi_random_2007} approximating an RBF kernel ($\gamma=1/64$),
followed by balanced logistic regression.

\textbf{Random forest.} \emph{Benchmark:} \num{100} trees, \texttt{max\_depth}$=16$,
\texttt{min\_samples\_leaf}$=50$, balanced class weights, on the common subsample.
\emph{Tuned (ensemble member):} \num{80} trees, \texttt{max\_features}$=12$,
\texttt{min\_samples\_leaf}$=14$, full bootstrap samples, Gini criterion, balanced class weights, on the full training
split, chosen by a sweep over feature subsampling, leaf size and tree count without any size limit.

\textbf{Histogram gradient boosting.} Gradient-boosted decision trees
(\texttt{max\_iter}$=300$, learning rate \num{0.1}, balanced weights)---the most
accurate single tree model on the common subsample.

\textbf{MLP.} \emph{Benchmark:} two hidden layers ($128\!\to\!64$), ReLU activations and early stopping,
trained on standardised embeddings. \emph{Tuned (ensemble member):} two hidden layers of \num{256} ReLU units, no dropout,
standardised inputs, class-balanced binary cross-entropy, AdamW (learning rate $3\times10^{-3}$, weight decay $10^{-5}$),
batch size \num{16384} and a one-cycle learning-rate schedule over eight epochs, stopped after the first epoch (the epoch
chosen on the selection half); the prediction is the mean of five seeds. The configuration was selected by a random search
over \num{64} configurations \cite{bergstra_random_2012} (depth and width, dropout, weight decay, learning rate, batch size,
schedule length and warm-up, input standardisation), scoring every configuration as the RF's partner in the ensemble.

\textbf{RF+MLP ensemble (selected).} Soft voting: $P(\text{solar}\mid\mathbf{x}) = \tfrac12\,P_{\text{RF}}(\mathbf{x}) + \tfrac12\,P_{\text{MLP}}(\mathbf{x})$
\cite{kuncheva_combining_2004}. All selection decisions (MLP configuration, stopping epoch,
finalist, ensemble weight) were made on one half of the country validation sites, split by \ang{0.5} cells, under a rule
fixed before the confirmation results were seen: ties within \num{0.001} ROC-AUC go to the cheaper model, and the weight
moves from $\tfrac12$ only for a gain of at least \num{0.001} (none did). Learned stacking and rank averaging were tested, but performed identically.

\section{Additional site-scale examples}
\label{app:examples}

\begin{figure}[H]
  \centering
  \includegraphics[width=\textwidth]{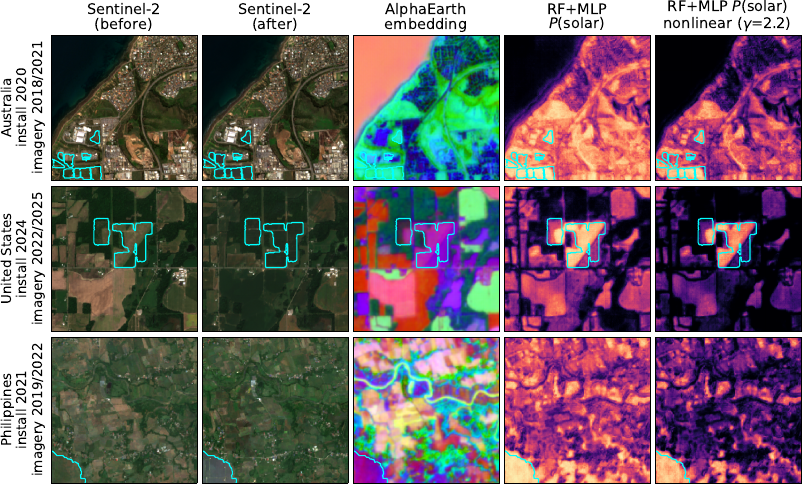}
  \caption{Site-scale suitability for three further installations (rows: Australia, United States,
  Philippines). Columns: Sentinel-2 before and after construction, the AEF embedding two years
  before installation, and the RF+MLP suitability map on a linear and a nonlinear ($\gamma=\num{2.2}$)
  colour scale; footprints outlined in cyan. Across all three geographies the embedding scores the
  future-solar parcels as suitable before any panels appear, and the nonlinear scale highlights the
  high-suitability cores.}
  \label{fig:examples}
\end{figure}

Fig.~\ref{fig:examples} complements the single site of Fig.~\ref{fig:rf}a with three
further installations spanning an urban coastal fringe (Australia), an agricultural interior
(United States) and a tropical smallholder landscape (Philippines). For each we show the
Sentinel-2 scene before and after construction, the AEF embedding two years before installation,
and the per-pixel RF+MLP suitability on both a linear scale and a nonlinear ($\gamma=\num{2.2}$)
rescaling that expands the high-suitability end of the range on an absolute \numrange{0}{1} axis. 
In every case the pre-installation embedding of the land under the footprint (cyan) is scored as
suitable; the uniformly-bright Philippines map illustrates the need for nonlinear rescaling, an issue also discussed in Appendix~\ref{app:weak} for dim countries.

\section{Forward validation}
\label{app:validation}
\textbf{Metrics and percentile capture.} Beyond ROC-AUC and PR-AUC,
we summarise deployment behaviour with a percentile-capture test: each real site is
assigned the fraction of random national land it outscores, its within-country
suitability percentile, from which we report the share of sites in the most-suitable
\num{20}\%, \num{10}\% and \num{5}\% of land, over the combined GPVF and TZ-SAM validation set. A typical future site sits at the \num{95}th
percentile of national land, with \num{85}\%, \num{66}\% and \num{48}\%
of sites in the most-suitable \num{20}\%, \num{10}\% and \num{5}\% of
land respectively.

\begin{figure}[H]
  \centering
  \includegraphics[width=0.62\textwidth]{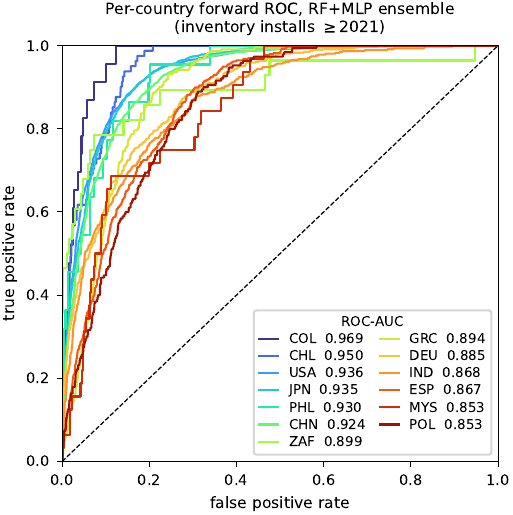}
  \caption{Per-country forward ROC curves for the RF+MLP ensemble on
  inventory sites installed from 2021 onward, scored against random national land.
  Step-like curves reflect the small validation counts of the data-poor countries
  (Table~\ref{tab:country}).}
  \label{fig:roc_curves}
\end{figure}

\textbf{Per-country results.} The ensemble separates real future sites from
random national land in all thirteen countries (Table~\ref{tab:country},
Fig.~\ref{fig:roc_curves}). The inventory and TZ-SAM sources agree closely (median
$|\Delta\text{ROC}|\approx\num{0.008}$), and pooling the far larger TZ-SAM positive
counts tightens the intervals for the data-poor countries by a median factor of \num{1.7}.
Countries are ordered by combined ROC-AUC.

\begin{table}[H]
  \centering
  \caption{Per-country forward validation of the RF+MLP ensemble, against inventory and independent TZ-SAM future sites. $n$ gives
  the number of validation sites; each is scored against random
  national land points. Bracketed ranges are bootstrap \num{95}\% confidence
  intervals.}
  \label{tab:country}
  \footnotesize
  \setlength{\tabcolsep}{5pt}
  \begin{tabular}{@{}l r l r c r l@{}}
    \toprule
    & \multicolumn{2}{c}{Inventory} & \multicolumn{2}{c}{TZ-SAM} & \multicolumn{2}{c}{Combined} \\
    \cmidrule(lr){2-3}\cmidrule(lr){4-5}\cmidrule(lr){6-7}
    Country & $n$ & ROC-AUC {[}\num{95}\% CI{]} & $n$ & ROC-AUC & $n$ & ROC-AUC {[}\num{95}\% CI{]} \\
    \midrule
    CHL & \num{81} & 0.950~[0.938, 0.962] & \num{163} & 0.955 & \num{244} & 0.953~[0.941, 0.964] \\
    COL & \num{23} & 0.969~[0.955, 0.982] & \num{93} & 0.948 & \num{116} & 0.952~[0.939, 0.963] \\
    JPN & \num{1450} & 0.935~[0.929, 0.941] & \num{517} & 0.929 & \num{1967} & 0.934~[0.928, 0.939] \\
    PHL & \num{22} & 0.930~[0.893, 0.964] & \num{51} & 0.924 & \num{73} & 0.926~[0.903, 0.946] \\
    USA & \num{2993} & 0.936~[0.931, 0.941] & \num{2200} & 0.903 & \num{5193} & 0.922~[0.917, 0.927] \\
    CHN & \num{2978} & 0.924~[0.918, 0.929] & \num{11084} & 0.903 & \num{14062} & 0.907~[0.902, 0.912] \\
    MYS & \num{32} & 0.853~[0.801, 0.898] & \num{48} & 0.932 & \num{80} & 0.900~[0.871, 0.929] \\
    GRC & \num{221} & 0.894~[0.880, 0.906] & \num{655} & 0.894 & \num{876} & 0.894~[0.885, 0.901] \\
    ZAF & \num{28} & 0.899~[0.819, 0.965] & \num{140} & 0.891 & \num{168} & 0.893~[0.864, 0.918] \\
    ESP & \num{397} & 0.867~[0.853, 0.879] & \num{817} & 0.864 & \num{1214} & 0.865~[0.856, 0.874] \\
    IND & \num{464} & 0.868~[0.853, 0.884] & \num{2086} & 0.860 & \num{2550} & 0.862~[0.854, 0.870] \\
    DEU & \num{465} & 0.885~[0.871, 0.899] & \num{952} & 0.846 & \num{1417} & 0.858~[0.849, 0.868] \\
    POL & \num{190} & 0.853~[0.832, 0.872] & \num{1042} & 0.821 & \num{1232} & 0.826~[0.816, 0.837] \\
    \bottomrule
  \end{tabular}
\end{table}

\textbf{Further validation.} We validate the ensemble along three axes, each
isolating a bias source. \emph{(i)~Forward in time:} each national map uses the 2019
embedding while validation sites were installed from 2021 onward, and any validation site belonging to a training-split
site is removed. \emph{(ii)~Across space} (leave-one-country-out, LOCO): the target
country is withheld from training, testing transfer to unseen geographies.
\emph{(iii)~Leak-proof temporal:} training is restricted to imagery up to 2021 (installs up to 2023) across all countries and
validated on 2024 installs, on the 2022 embedding. Every country is additionally cross-checked against
the independent TZ-SAM inventory, a different source absent from training, with
bootstrap \num{95}\% confidence intervals. The LOCO and temporal models are
deliberately handicapped diagnostics (the same RF+MLP recipe refit without the withheld country or years) that bound
the two leakage pathways; all other results use the ensemble trained on the \num{80}\% training split, and the released
layer uses the same recipe refit on all data.

\begin{figure}[H]
  \centering
  \includegraphics[width=\textwidth]{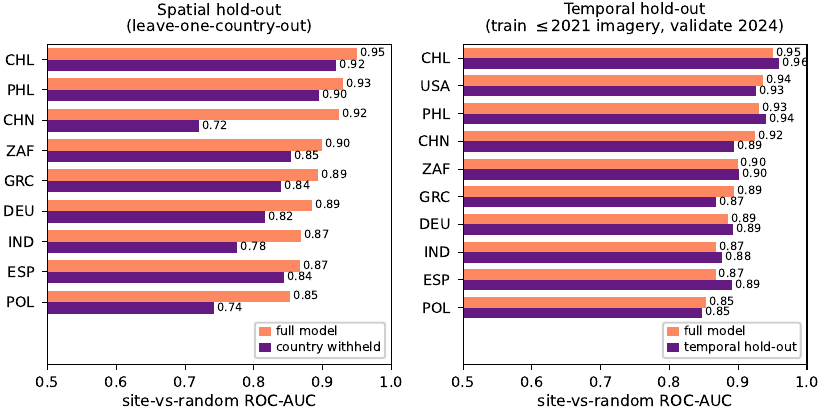}
  \caption{Site-versus-random ROC-AUC of the RF+MLP ensemble (full model) against the same recipe refit as
  hold-out models, per country. \emph{Left:} leave-one-country-out spatial hold-out. \emph{Right:}
  leak-proof temporal hold-out (training imagery $\leq$2021, 2024 installs; countries with a forward set).}
  \label{fig:ablations}
\end{figure}

Fig.~\ref{fig:ablations} outlines both LOCO and temporal ablation results. For the spatial holdout (nine countries), performance falls furthest for the
hardest, most distinctive countries (China $0.92\!\to\!0.72$, Poland $0.85\!\to\!0.74$, India
$0.87\!\to\!0.78$) and least for the easier ones (Spain, Philippines, Chile), averaging
\num{0.82}, so the optimism premium tracks country difficulty rather than being
uniform. The temporal hold-out (twelve countries) is
fully leak-proof in time; it stays strong (mean \num{0.90}, pooled global-2024
\num{0.89}) and matches or exceeds the in-sample forward number in six of the ten countries with both (e.g.\ India,
Spain, Germany), confirming the forward result is not a temporal-leakage artefact.

\section{Weak-country analysis}
\label{app:weak}

\begin{figure}[H]
  \centering
  \includegraphics[width=\textwidth]{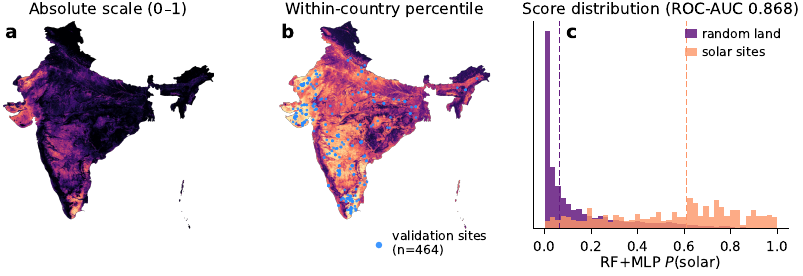}
  \caption{\textbf{(a)}~The RF+MLP suitability field on an absolute \numrange{0}{1} scale: India's
  land scores stay low, so the country map looks dim. \textbf{(b)}~Re-expressing
  the same field as within-country percentiles brightens it and reveals structure;
  overlaid validation sites (blue) concentrate in the high-percentile regions.
  \textbf{(c)}~Score histogram for random land versus solar sites
  (ROC-AUC \num{0.868}); the percentile transform is rank-preserving, so it leaves
  ROC-AUC unchanged.}
  \label{fig:weak}
\end{figure}

Performance ranges from ROC-AUC \num{0.83} (Poland) to \num{0.95} (Chile), but this
spread reflects country difficulty, not training-data volume: per-country ROC
correlates weakly negatively with the number of sites (Spearman $\rho=\num{-0.26}$; China, with
\num{14000}$+$ sites, scores \num{0.91}; Chile, with \num{244}, scores \num{0.95}),
so the harder countries are geophysically diverse rather than data-starved. The effect is one of readability, not skill. In a dim country such as India the
raw scores are compressed into a narrow band (Fig.~\ref{fig:weak}a), so
re-expressing each country's land as percentiles of its own score distribution
brightens the map (Fig.~\ref{fig:weak}b) while leaving ROC-AUC unchanged at
\num{0.868}. This makes the percentile view the more readable
one for national maps. The raw scores nonetheless share one global scale and each country's future sites outrank the
random land of the other twelve countries (ROC-AUC \num{0.83}--\num{0.96}).

\section{Discrimination by installed capacity}
\label{app:size}

\begin{figure}[H]
  \centering
  \includegraphics[width=0.65\textwidth]{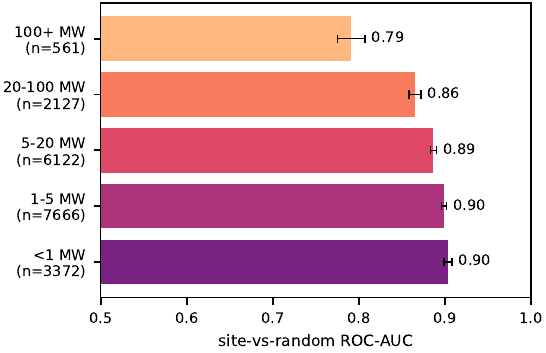}
  \caption{Site-versus-random ROC-AUC of the RF+MLP ensemble by installed capacity (bootstrap \num{95}\% CIs): strong for the small-to-mid sites that dominate deployment, declining for the rare $>$\SI{100}{MW} utility farms.}
  \label{fig:sizeroc}
\end{figure}

We bin the combined GPVF$+$TZ-SAM validation sites by their TZ-SAM installed-capacity estimate and compute the site-versus-random ROC-AUC within each capacity class (Fig.~\ref{fig:sizeroc}). Discrimination is strong across the small-to-mid solar farms that dominate deployment (ROC-AUC \num{0.90} below \SI{5}{MW}) and falls for the rarer, atypically-sited utility farms above \SI{100}{MW} (\num{0.79}), whose siting is often driven by land availability and heavy transmission infrastructure rather than the qualities the model has learnt. Because the model scores pixels, it does not test whether enough contiguous suitable land exists for a plant of a given size.

\section{Covariate analysis}
\label{app:covariate}

\begin{figure}[H]
  \centering
  \includegraphics[width=\textwidth]{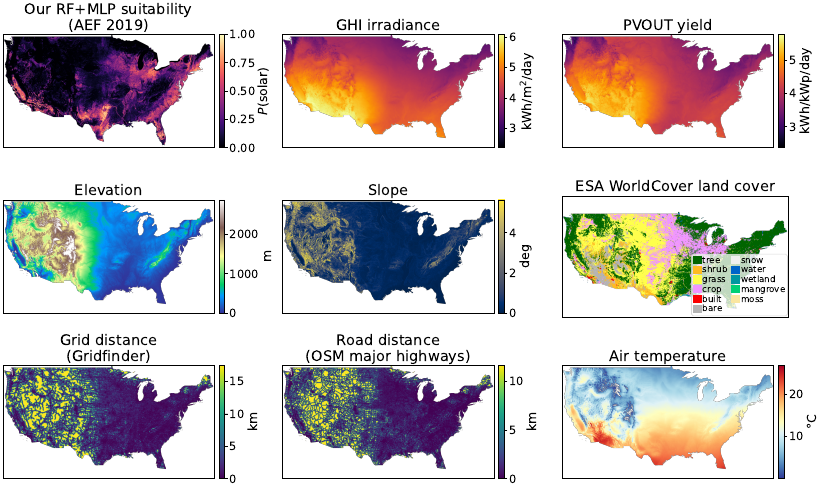}
  \caption{The physical-covariate stack for the United States alongside our RF+MLP
  suitability (top-left, AEF-2019, raw $P$(solar)). Suitability is high across the
  developed central and eastern land and low over the remote mountain west, tracking land
  cover, terrain and infrastructure (grid- and road-distance) rather than the irradiance
  layers (GHI, PVOUT), which peak in the south-west where comparatively little solar is
  actually built.}
  \label{fig:uscov}
\end{figure}

\textbf{Method.} To test whether the embedding carries siting information beyond
standard GIS layers, we build a covariate stack for each country: land cover (ESA WorldCover 2021 \citep{zanaga_esa_2022}), slope and elevation (Copernicus GLO-30 DEM \citep{copernicus_dem_2021}), air temperature, the irradiance components
GHI (global horizontal irradiance), GTI (global tilted irradiance), DNI (direct normal irradiance) and DIF (diffuse horizontal irradiance), and long-term PVOUT (photovoltaic output), all from the Global Solar Atlas \citep{global_solar_atlas}; distance to roads (OpenStreetMap \citep{openstreetmap}) and to the electricity grid (Gridfinder \citep{arderne_predictive_2020}); population density (GPWv4.11 \citep{ciesin_gpw_2018}); and protected-area
status (WDPA \citep{wdpa}). We regress the ensemble's suitability score on the covariate stack with both a linear model and a
random forest. The residual that the covariate model cannot reconstruct ($1-R^2$) measures
the information the embedding holds that these layers do not; per-covariate
permutation importance reveals which physical factors the learned suitability aligns
with. Fig.~\ref{fig:uscov} shows part of this stack for the United States.

\begin{figure}[H]
  \centering
  \includegraphics[width=\textwidth]{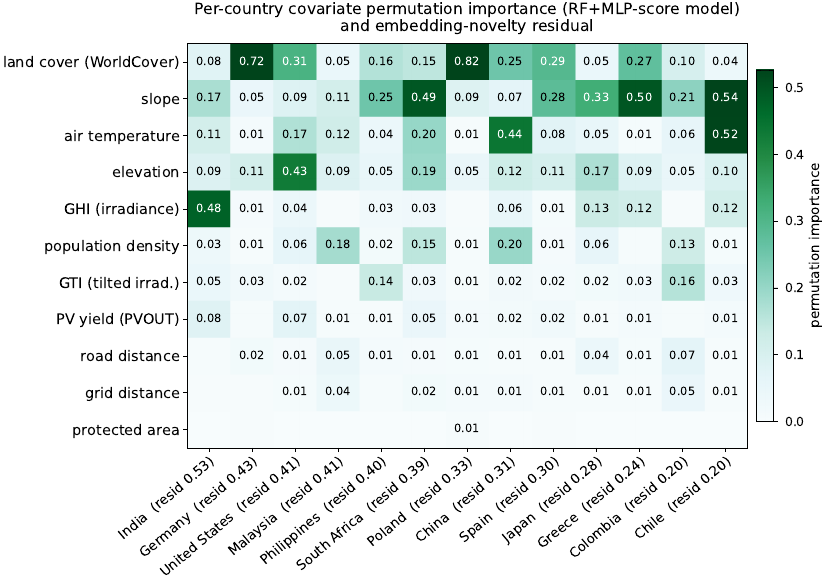}
  \caption{Per-country covariate permutation importance of a model
  predicting the RF+MLP ensemble suitability score from GIS layers, with the embedding-novelty
  residual under each country (higher $=$ more the embedding knows that the layers
  do not). WorldCover classes are summed into a single land-cover row.}
  \label{fig:importance}
\end{figure}

\textbf{Result.} The residual runs from \num{0.20} (Chile) to \num{0.53}
(India), and the random-forest fit consistently exceeds the linear one, confirming the information is non-linear. The factors the suitability aligns with most are land cover and terrain (slope, elevation) and air temperature. Infrastructure and irradiance (GHI/GTI/PVOUT)
matter only in a few resource-driven countries and are rarely dominant (Fig.~\ref{fig:importance}).

\textbf{Orthogonality to solar resource products.} The learned suitability is essentially
orthogonal to the authoritative long-term photovoltaic output potential (PVOUT)
product \cite{noauthor_global_nodate}: across countries their rank correlation is near
zero or weakly negative (Spearman $\rho\approx\num{-0.10}$), and PVOUT alone cannot separate real sites from
random land (ROC-AUC \num{0.49} versus our \num{0.90}). Conventional siting anchored
on such resource layers therefore misranks deployment, as the China comparison
(Fig.~\ref{fig:china}) shows.

\section{MCDA benchmark}
\label{app:mcda}

\textbf{MCDA construction.} We reconstruct multi-criteria suitability from two published expert weightings: an infrastructure-led scheme \citep{richards_sustainable_2025} and a resource-and-terrain scheme \citep{chen_jong_multi-criteria_2024}, listed in Table~\ref{tab:mcda_weights}. Criteria are built from the same data layers as the covariate analysis (Appendix~\ref{app:covariate}), with aspect derived from the DEM. Each criterion is scaled to $[0,1]$ by country-relative 2--98\% min--max normalisation.

\begin{table}[H]
  \centering
  \caption{Published expert criterion weights reconstructed for the MCDA benchmark, from \citet{richards_sustainable_2025} and \citet{chen_jong_multi-criteria_2024}; a dash marks a criterion the scheme omits and each column sums to one. GHI/GTI: global horizontal/tilted irradiance; PVOUT: long-term photovoltaic output potential.}
  \label{tab:mcda_weights}
  \small
  \setlength{\tabcolsep}{10pt}
  \begin{tabular}{@{}l c c@{}}
    \toprule
    Criterion & Richards et al.\ & Chen et al.\ \\
    \midrule
    Road distance      & 0.54 & --   \\
    Grid distance      & 0.14 & 0.17 \\
    Air temperature    & 0.08 & 0.17 \\
    Aspect             & 0.08 & --   \\
    Slope              & 0.06 & 0.16 \\
    GHI                & 0.04 & 0.19 \\
    PVOUT              & 0.03 & --   \\
    Elevation          & 0.02 & 0.16 \\
    GTI                & 0.02 & --   \\
    Population density & --   & 0.15 \\
    \bottomrule
  \end{tabular}
\end{table}

\textbf{Aggregation and land constraints.} We combine the weighted criteria with two standard aggregators. \emph{Weighted linear combination} (WLC) is the weighted sum of the normalised criteria. \emph{TOPSIS} (Technique for Order of Preference by Similarity to Ideal Solution) instead ranks each location by its relative closeness to a fixed ideal point that is best on every criterion (all ones) against an anti-ideal that is worst on every criterion (all zeros): the score is $d^{-}/(d^{+}+d^{-})$, with $d^{+}$ and $d^{-}$ the Euclidean distances to the ideal and anti-ideal, so a location ranks high when it lies near the ideal and far from the anti-ideal. Land cover and protected areas enter in one of two ways: either as a hard exclusion only---WDPA protected areas and unsuitable WorldCover classes (open water, built-up, permanent ice) forced to zero---or, additionally, as a continuous weighted land-suitability criterion layered on top of that exclusion. Crossing the two weight sets, the two aggregators, and the two land treatments yields eight published-weight MCDA configurations per country; we add their equal-weight counterparts (each scheme's criteria weighted uniformly), giving \textbf{sixteen} configurations. Each is scored on the same combined GPVF$+$TZ-SAM site-versus-random task as the ensemble (paired bootstrap over sites), and we report the strongest, a deliberately favourable baseline for MCDA.

\begin{table}[t]
  \centering
  \caption{RF+MLP ensemble versus published MCDA, combined GPVF$+$TZ-SAM sites, ordered by gap. MCDA is the best of sixteen
  configurations (Richards/Chen weights and their equal-weight variants $\times$ WLC/TOPSIS $\times$ exclusion/weighted land). Bracketed ranges are
  paired bootstrap \num{95}\% confidence intervals on the gap; all lie
  strictly above zero.}
  \label{tab:mcda}
  \small
  \setlength{\tabcolsep}{6pt}
  \begin{tabular}{@{}l r r r l@{}}
    \toprule
    Country & $n$ & RF+MLP & MCDA & Gap {[}\num{95}\% CI{]} \\
    \midrule
    China & \num{27158} & 0.914 & 0.615 & \textbf{+0.299}~[0.291, 0.307] \\
    Chile & \num{544} & 0.961 & 0.725 & \textbf{+0.237}~[0.218, 0.256] \\
    Greece & \num{1735} & 0.916 & 0.683 & \textbf{+0.233}~[0.218, 0.249] \\
    United States & \num{7564} & 0.929 & 0.741 & \textbf{+0.189}~[0.180, 0.197] \\
    India & \num{4386} & 0.886 & 0.712 & \textbf{+0.174}~[0.165, 0.184] \\
    Poland & \num{2062} & 0.845 & 0.671 & \textbf{+0.174}~[0.161, 0.187] \\
    Spain & \num{2775} & 0.894 & 0.720 & \textbf{+0.174}~[0.163, 0.186] \\
    Japan & \num{3246} & 0.941 & 0.779 & \textbf{+0.161}~[0.151, 0.172] \\
    Malaysia & \num{163} & 0.937 & 0.788 & \textbf{+0.149}~[0.107, 0.197] \\
    Germany & \num{2864} & 0.886 & 0.740 & \textbf{+0.146}~[0.135, 0.157] \\
    Philippines & \num{167} & 0.928 & 0.831 & \textbf{+0.097}~[0.070, 0.125] \\
    South Africa & \num{270} & 0.921 & 0.828 & \textbf{+0.093}~[0.070, 0.116] \\
    Colombia & \num{198} & 0.963 & 0.905 & \textbf{+0.058}~[0.041, 0.078] \\
    \bottomrule
  \end{tabular}
\end{table}

\begin{figure}[b]
  \centering
  \includegraphics[width=0.88\textwidth]{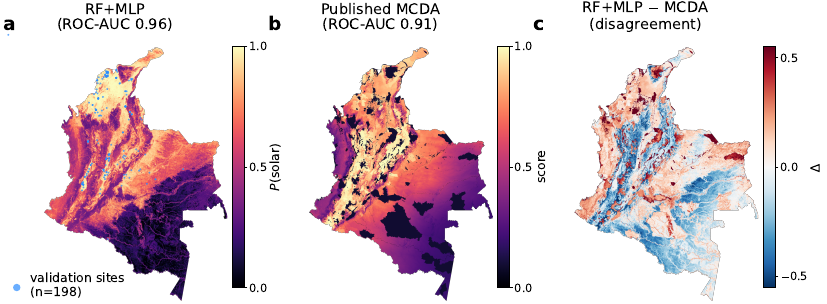}
  \caption{Colombia RF+MLP-versus-MCDA maps, the analog of Fig.~\ref{fig:china} for a
  resource-driven country. \textbf{(a)}~RF+MLP suitability (AEF-2019) with the \num{198}
  scored validation sites (blue), concentrated in the Andean populated corridor (ROC-AUC
  \num{0.96}). \textbf{(b)}~Best published MCDA (Chen weights, GHI$+$population; ROC-AUC
  \num{0.91}). \textbf{(c)}~RF+MLP$-$MCDA disagreement. Unlike China, the ensemble and MCDA largely
  agree here so the ensemble's edge is small (${+}\num{0.058}$),
  illustrating that the advantage concentrates where expert weights misread the geography. Certain areas of disagreement, particularly protected areas, are still visible.}
  \label{fig:colombia}
\end{figure}

\textbf{Result.} The ensemble beats the best MCDA configuration in all thirteen
countries, and every gap is significant (Table~\ref{tab:mcda}). The advantage is
largest exactly where expert weights misread the geography, China (${+}\num{0.30}$),
where MCDA over-weights the irradiance that anti-correlates with deployment, and
smallest where siting genuinely tracks the resource (Colombia, South Africa).
Fig.~\ref{fig:colombia} shows the Colombia case (the analog of Fig.~\ref{fig:china}):
a solar resource driven country where the two largely agree.

\end{document}